%% file: main.tex
\documentclass[sigconf]{acmart}
\AtBeginDocument{%
  }

\setcopyright{none}
\usepackage{amsmath}
\usepackage{algorithm}
\usepackage{algpseudocode}
\begin{document}

\title{Towards Standardized Evaluation in Automated Domain Modeling: Introducing a Benchmark}

\author{Vasiliy Seibert}
\affiliation{%
  \institution{Institute for Software and Systems Engineering, TU Clausthal}
  \city{Clausthal-Zellerfeld}
  \state{Lower Saxony}
  \country{Germany}
}
\email{vasiliy.seibert@tu-clausthal.de}
\orcid{0000-0002-7121-6816}

\begin{abstract}
Domain modeling plays an essential role in domain-driven design, capturing essential entities and their relationships within a specific domain. Despite advancements in automated domain modeling, the absence of standardized benchmarks has hindered the comparative assessment of existing approaches. This paper introduces a benchmark designed to address this gap. The benchmark combines the 45-record Golden UML Modelset (Verbruggen et al., 2025) on Zenodo, as distributed by the Text2UML project of Calamo, Mecella, and Snoeck (Calamo et al., 2025), with the 8-record reference archive of Chen et al. (Chen et al., 2023a,b), enabling the evaluation of automated domain modeling approaches across different levels of complexity and scale. Given a natural language description, the task is to generate a corresponding domain model. For each description, a reference domain model is provided as ground truth. A metric is used to compare the generated domain model with the corresponding ground-truth model. To demonstrate the utility of the benchmark, we evaluate multiple automated domain modeling approaches, including heuristic rule-based methods and LLM-driven strategies. In accordance with the FAIR4RS recommendations (Chue Hong et al., 2022), the benchmark is provided as a research artifact to encourage reuse and support future research on automated domain modeling.
\end{abstract}

\ccsdesc[500]{Software and its engineering → Software creation and management → Software verification and validation}

\keywords{domain modeling, class diagrams, large language models, benchmark, evaluation, FAIR principles}

\received{2026-07-03}

\maketitle

\input{chapters/01_introduction.tex}
\input{chapters/02_related_work.tex}

\input{chapters/03_approach.tex}

\input{chapters/04_results.tex}

\bibliographystyle{ACM-Reference-Format}
\bibliography{main}

\end{document}

%% file: chapters/01_introduction.tex

\section{Introduction}
\label{sec:01_introduction}


Domain modeling is central to model-driven software engineering~\cite{evans2004ddd}, and a domain model captures the relevant entities of a domain and their relationships using the syntax of {UML} class diagrams. The benefit of using {UML} class-diagram syntax is that the resulting model is precise and well known by software developers, who will use it as a foundation for building the system, and that the model facilitates clear communication in requirements discussions and reduces misunderstandings between stakeholders. \newline


Assisting or even automating domain modelling has great potential regarding increasing quality and reducing cost. Traditionally, heuristic rule-based methods~\cite{ahmed2022automatic,abdelnabi2020generating} have been formulated to analyze Requirement Descriptions given in Natural Language Text using tools like the Stanford CoreNLP API and subsequently apply rules to extract entities and meaningful relationships between them. However, these approaches are getting surpassed by modern approaches that rely entirely on Large Language Models ({LLMs})~\cite{camaracamara2023chatgpt}. \newline


Measuring the performance of automated domain modelling approaches has been neglected in the literature~\cite{camaracamara2023chatgpt}. A benchmark analogous to {Stanford}'s {HELM} framework~\cite{helm2024stanford}, consisting of a clearly defined task, a public dataset, and a deterministic metric, would help the research community by providing a place for healthy competition. The literature has repeatedly called for such a benchmark: C\'{a}mara et al.~\cite{camaracamara2023chatgpt} state that the assessment of generative {AI} in modeling tasks is under-investigated, and C\'{a}mara, Burgue\~{n}o, and Troya~\cite{camara2024towards} argue for robust methodologies to standardize it. On the dataset side, a community-curated domain model corpus~\cite{verbruggen2025golden,calamo2025text2uml} has emerged as a common reference set. On the metric side, the Seibert study~\cite{seibert2025metrik} implements five metrics behind a single interface and evaluates them against human expert ratings~\cite{chen2023automated}. \newline


The contribution of this paper is to address the previously described gap by proposing a benchmark. The task is to take a natural-language description of a domain~\cite{evans2004ddd} and to generate a domain model in {PlantUML} class-diagram notation. The benchmark uses a dataset of 53 domain models with their corresponding natural-language descriptions, drawn from two sources~\cite{verbruggen2025golden,calamo2025text2uml,chen2023automated,chen2023zenodo}. The metric is taken from the Seibert study~\cite{seibert2025metrik}, which implements five metrics from four source papers~\cite{singh2022detecting,cech2019matching,yuan2020structural,triandini2021automated} behind a single interface, each returning three sub-scores in the unit interval, one per element type (class, attribute, association), where each sub-score is the {F1} of the element-level match between candidate and reference. The benchmark uses metrik-4, the Triandini~\cite{triandini2021automated} metric, on the basis that the Seibert study~\cite{seibert2025metrik} shows it to be the strongest of the five on per-element agreement with the Chen et al.~\cite{chen2023automated} human expert ratings, winning 2 of 3 \mbox{element-by-statistic} cells. \newline


To encourage reuse, the benchmark is released as a research artifact in accordance with the {FAIR4RS} recommendations~\cite{chuehong2022fair4rs, wilkinson2016fair} \url{https://doi.org/10.5281/zenodo.21166386}. \newline

%% file: chapters/02_related_work.tex

\section{Related Work}
\label{sec:02_related_work}


Prior approaches to automated domain modeling each evaluate on their own dataset, and the field's adoption of a common reference set is recent. Heuristic rule-based methods~\cite{clavel2007mova,popescu2008reducing,bajwa2009object,vinay2009approach,sharma2009extracting,bajwa2012natural,arora2016extracting,benabdessalemkaraa2016abcd,mohanan2018natural,abdelnabi2020generating} each report a private, case-specific, or industrial evaluation, with no public artefact deposited at a stable identifier. Chen et al.~\cite{chen2023automated} compare different {LLM}s and prompting strategies, bring their own dataset, and grade candidate models through human expert comparisons despite automated metrics being available, though they do deposit their data on {Zenodo}~\cite{chen2023zenodo}. Parts of the Chen data are incorporated in a community-curated domain model corpus~\cite{verbruggen2025golden}, published at a stable identifier with peer-reviewed provenance. The Calamo, Mecella, and Snoeck corpus~\cite{calamo2025text2uml,calamo2026text2umlresults} is a cleaned encoding of the Verbruggen 45 published alongside a generation pipeline, and is, to our knowledge, the first reference corpus in this space to be published with the pipeline that consumes it. \newline


Prior approaches to grading a candidate domain model against a reference are diverse in approach, and no metric has become a community standard. In the education application area, Modi, Taher, and Mahmud~\cite{modi2021tool} develop a Java-based tool that produces an automatic grade from student and instructor solution diagrams, Boubekeur, Mussbacher, and McIntosh~\cite{boubekeur2020automatic} combine heuristics with machine learning to predict approximate letter grades, and Bouali et al.~\cite{bouali2025llmgrading} compare {LLM}-generated scores against teaching assistants on 92 student submissions. Singh, Boubekeur, and Mussbacher~\cite{singh2022detecting} propose a rule-based Mistake Detection System covering 83 of 97 identified mistake types, and Fauzan, Siahaan, Rochimah, and Triandini~\cite{fauzan2021automated,triandini2021automated} decompose assessment into a structural and a semantic component, reporting agreement with expert judgements. For model reuse and repository mining, \v{C}ech~\cite{cech2019matching} defines a class-model distance via graph edit distance, Yuan, Yan, and Ma~\cite{yuan2020structural} decompose structural similarity into intra- and inter-structure components over a {UML} Class Graph, and Song et al.~\cite{song2024deep} adopt a graph neural network on use case graphs with {TF-IDF} cosine similarity. None of these approaches is cross-compared with a metric from another family, and each evaluates on its own dataset. The Seibert study~\cite{seibert2025metrik} implements five metrics from four source papers~\cite{singh2022detecting,cech2019matching,yuan2020structural,triandini2021automated} behind a single interface and compares them against human expert ratings~\cite{chen2023automated} on a common corpus. This benchmark reuses one of the five implementations, namely metrik-4, the Triandini~\cite{triandini2021automated} metric, as the fixed metric. \newline


%% file: chapters/03_approach.tex

\section{Approach}
\label{sec:03_approach}

\subsection{Benchmark Approach}

The benchmark decomposes into three axes: a data axis, a task axis, and a metric axis. The data axis holds 53 records, each a pair of a natural-language text and a reference {PlantUML} class diagram drawn from two community-curated corpora. The task axis runs each candidate strategy against the natural-language text to produce a generated {PlantUML} class diagram. The metric axis is a deterministic comparator that accepts the reference and generated diagrams and returns a 3-tuple of per-element similarity scores, one each for classes, attributes, and associations. \newline

\subsection{Data}

\begin{figure*}[htbp]
\centering
\includegraphics[width=0.75\textwidth]{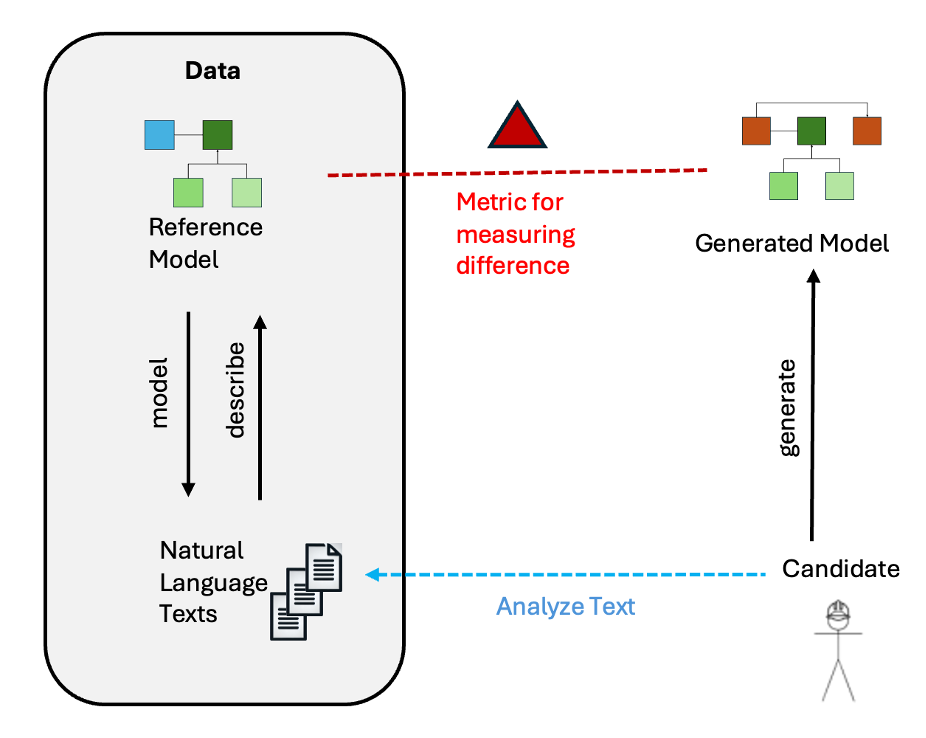}
\caption{Overview of the benchmark: a (natural-language text, reference {PlantUML} class diagram) pair is fed to a candidate strategy, which produces a candidate {PlantUML} diagram that a metric compares against the reference.}
\label{fig:3-1}
\end{figure*}


The benchmark runs on 53 records drawn from two {PlantUML} class-diagram corpora: a 45-record community-curated corpus~\cite{verbruggen2025golden,calamo2025text2uml} and an 8-record reference archive~\cite{chen2023automated,chen2023zenodo}. The eight reference diagrams were derived from the ground truth of the Chen et al. study; while that study produced many more models, those are generated by {LLM}s, whereas these eight were crafted by human experts in the field of education. Each record is a pair of a natural-language text and a reference {PlantUML} class diagram that the text describes. Together, the 53 records contribute 540 classes, 667 attributes, 184 enums, and 710 relationships. The two corpora differ in scale and in per-record composition: the 45-record corpus averages 9.4 classes and 13.0 attributes per record, while the 8-record archive averages 14.4 classes and 10.3 attributes, reflecting the denser modelling style of its source domain. Table~\ref{tab:data-overview} breaks down the 53 records per dataset and decomposes the 710 relationships into seven relationship types included in the dataset, ranging from 355 plain associations to 2 association-class links. Across both corpora, 70\% of relationship endpoints carry a multiplicity, indicating that the majority of relationships specify cardinality information that a candidate model must recover. \newline

\begin{table}[t]
\centering
\small
\begin{tabular}{lrrr}
\hline
 & DS-1 (45) & DS-2 (8) & Total (53) \\
\hline
NLT words (mean) & 316 & 423 & --- \\
Classes (mean / total) & 9.4 / 425 & 14.4 / 115 & 540 \\
Attributes (mean / total) & 13.0 / 585 & 10.3 / 82 & 667 \\
Enums (mean / total) & 2.4 / 109 & 9.4 / 75 & 184 \\
Relationships (mean / total) & 11.9 / 537 & 21.6 / 173 & 710 \\
Association (\verb|--|) & 268 & 87 & 355 \\
Aggregation (\verb|o--|) & 6 & 0 & 6 \\
Inheritance (\verb|--|>|) & 102 & 34 & 136 \\
Composition (\verb|*--|) & 116 & 35 & 151 \\
Directed assoc.\ (\verb|-->|) & 32 & 1 & 33 \\
Dependency (\verb|..|) & 13 & 14 & 27 \\
Association-class & 0 & 2 & 2 \\
Card (\% with multiplicity) & 68 & 76 & 70 \\
\hline
\end{tabular}
\caption{Per-dataset overview.}
\label{tab:data-overview}
\end{table}

\noindent\textbf{LabTracker natural-language text~\cite{chen2023automated,chen2023zenodo}:}
\begin{small}
\begin{verbatim}
The LabTracker software helps (i) doctors manage the
requisition of tests and examinations for patients and
(ii) patients book appointments for tests and
examinations at a lab. For the remainder of this
description, tests and examinations are used
interchangeably. For a requisition, a doctor must
provide their numeric practitioner number and
signature for verification as well as their full name,
their address, and their phone number. The signature
is a digital signature, i.e., an image of the actual
signature of the doctor. Furthermore, the doctor
indicates the date from which the requisition is
valid. The requisition must also show the patient's
information including their alpha-numeric health
number, first name and last name, date of birth,
address, and phone number. A doctor cannot prescribe
a test for themselves but can prescribe tests to
someone else who is a doctor. Several tests can be
combined on one requisition but only if they belong
to the same group of tests. For example, only blood
tests can be combined on one requisition or only
ultrasound examinations can be combined. It is not
possible to have a blood test and an ultrasound
examination on the same requisition. For each test,
its duration is defined by the lab network, so that
it is possible to schedule appointments accordingly.
The duration of a test is the same at each lab. For
some kinds of tests, it does not matter how many
tests are performed. They take as long as a single
test. For example, several blood tests can be
performed on a blood sample, i.e., it takes as long
to draw the blood sample for a single blood test as
it does for several blood tests. A doctor may also
indicate that the tests on a requisition are to be
repeated for a specified number of times and
interval. The interval is either weekly, monthly,
every half year, or yearly. All tests on a
requisition are following the same repetition
pattern. The doctor and the patient can view the
results of each test (either negative or positive)
as well as the accompanying report. A patient is
required to make an appointment for some tests while
others are walk-in only. For example, x-ray
examinations require an appointment, but blood tests
are walk-in only (i.e., it is not possible to make
an appointment for a blood test). On the other hand,
some tests only require a sample to be dropped off
(e.g., a urine or stool sample). To make an
appointment for a requisition, a patient selects the
desired lab based on the lab's address and business
hours. For requisitions with repeated tests, a
patient is only allowed to make one appointment at a
time. The confirmation for an appointment also shows
a confirmation number, the date as well as start/end
times, and the name of the lab as well as its
registration number. It is possible to change or
cancel an appointment at any time but doing so
within 24 hours of the appointment incurs a
change/cancellation fee. Each lab determines its own
fee and business hours. All labs are open every day
of the year and offer all tests. The business hours
of a lab do not change from one week to the next.
Each day a lab is open from the day's start time to
its end time, i.e., there are no breaks.
\end{verbatim}
\end{small}

\vspace{1em}

\noindent\textbf{LabTracker reference {PlantUML} (truncated)~\cite{chen2023automated,chen2023zenodo}:}
\begin{small}
\begin{verbatim}
@startuml
enum Interval {
  WEEKLY
  MONTHLY
  HALF_YEARLY
  YEARLY
}
class Doctor {
  +signature: string
}
class Patient {
  +dateOfBirth: string
}
class Requisition {
  +effectiveDate: string
  +repetitionInterval: Interval
}
(...)
Doctor "1" -- "*" Requisition
Requisition -- "*" Test
Patient "1" -- "0..*" Requisition
Appointment "0..*" -- "1" Lab
Test "1" -- "0..1" Appointment
@enduml
\end{verbatim}
\end{small}

\subsection{Metric}

This benchmark makes use of the metric implementations provided by the Seibert study~\cite{seibert2025metrik}, which packages five metrics from four source papers~\cite{singh2022detecting,cech2019matching,yuan2020structural,triandini2021automated} behind a single interface that accepts two {PlantUML} strings and returns a 3-tuple of per-element similarity scores, one each for classes, attributes, and associations. The study identifies two risks in reusing published metrics: a wrong implementation of the metric, and a wrong translation of the metric's native output format into the per-element {F1} scale of the human expert ratings. To address the first risk, each metric is restated as a \mbox{design-by-contract} specification with pre- and post-conditions, decomposed into sub-functions, and handed unchanged to two independent {LLM}s through the same harness; if both implementations produce identical output on all 39 ground-truth comparisons from Chen et al.~\cite{chen2023automated}, the specification is assumed to be well-defined and the implementations are assumed to be correct. To address the second risk, the specification is extended to translate each metric's native output into the per-element {F1} scale, and the translation is tested for partial-order preservation against the metric's native output, so that the ordering of the metric results is consistent with the ordering of the original output across the 39 comparisons. The five implementations are then compared against human expert ratings using a consistency table of four statistics --- per-element {MAD}, residual standard deviation, absolute bias, and Pearson r (linear correlation between metric score and human F1) --- computed across the 39 comparisons for each of the three element types. No single metric dominated across all cells: the per-element winners split across M-1, M-3, M-4, and M-5, with M-3 achieving the highest correlation on attributes (r = 0.65) and M-4 achieving the best correlation on classes (r = 0.42) and relationships (r = 0.42). A qualitative analysis complemented the statistical comparison, examining where each metric agreed with or diverged from the human judgements and identifying systematic biases such as M-4's constant overestimation on relationships. The benchmark chooses metrik-4, the Triandini~\cite{triandini2021automated} metric, because it wins 2 of the 3 \mbox{element-by-r} cells and also achieves the best attribute {MAD} (0.14), making it the most balanced metric on the consistency table. The implementation is reused from the upstream \texttt{\mbox{domain-model-metrics}} package (v1.0.0) without modification, so the per-element scores are reproducible from the version pin. \newline
\vspace{1em}

\noindent\textbf{Seibert study metric interface:}
\begin{small}
\begin{verbatim}
Metric M
  compute(ReferencePlantUML,
          GeneratedPlantUML)
    -> (class_score      : [0, 1],
       attribute_score   : [0, 1],
       association_score : [0, 1])
\end{verbatim}
\end{small}

\subsection{Candidates}

The benchmark reuses three open-source candidate strategies: one deterministic rule-based pipeline and two {LLM}-driven zero-shot strategies. The {rule\_based} candidate is a \mbox{Python-over-spaCy} re-implementation of the NLP-based pipeline of Abdelnabi, Maatuk, Abdelaziz, and Elakeili~\cite{abdelnabi2020generating}, which parses the natural-language text with spaCy's dependency parser and applies heuristic rules for extracting classes, attributes, and relationships, emitting a {PlantUML} class diagram with no {LLM} call. The two zero-shot candidates were derived from the literature: one from Chen et al.~\cite{chen2023automated,chen2023zenodo} and one from Calamo, Mecella, and Snoeck~\cite{calamo2025text2uml,calamo2026text2umlresults}. The Chen et al. prompt asks the {LLM} to extract the class diagram as an intermediate domain-specific language (Enumerations, Class, Relationships); the Calamo et al. prompt asks the {LLM} to emit the {PlantUML} class diagram directly, guided by a 5-step directive covering classes, attributes, relations and inheritance, \mbox{relation-to-class} assignment, and cardinalities. Each {LLM}-driven candidate is run with two {LLM}s ({GLM}-5.1 and {Kimi} {K2}) and repeated three times to account for the non-determinism of cloud-served {LLM}s. To ensure that the {LLM} output adheres to the {PlantUML} syntax accepted by the parser, the Chen et al. candidate makes a second {LLM} call prompted with a grammar encoding the parser's strict grammar, which translates the intermediate {DSL} into canonical {PlantUML}. \newline


\noindent\textbf{Chen et al.~\cite{chen2023automated,chen2023zenodo} --- Stage 1, system prompt:}
\begin{small}
\begin{verbatim}
Generate the lists of model classes and associations
from a given description.
\end{verbatim}
\end{small}

\vspace{1em}

\noindent\textbf{Chen et al.~\cite{chen2023automated,chen2023zenodo} --- Stage 1, task prompt:}
\begin{small}
\begin{verbatim}
Create a class diagram for the following description
by giving the enumerations, classes, and relationships
using format:
Enumerations:
enumerationName(literals)
(there might be no or multiple enumerations)

Class:
className(attributeType attributeName
(there might be multiple attributes))
(there might be multiple classes)

Relationships
mul1 class1 associate mul2 class2
(class1 and2 are classes above. mul1 and mul2
are one of the following options
[0..*, 1, 0..1, 1..*])
(there might be multiple associations)

class1 inherit class2
(class1 and class2 are classes above)
(there might be multiple inheritance)

mul1 class1 contain mul2 class2
(class1 and2 are classes above. mul1 and mul2
are one of the following options
[0..*, 1, 0..1, 1..*])
(there might be multiple composition)
\end{verbatim}
\end{small}

\vspace{1em}

\noindent\textbf{Chen et al.~\cite{chen2023automated,chen2023zenodo} --- Stage 2, translate prompt:}
\begin{small}
\begin{verbatim}
You are a PlantUML translator. You will be given a
draft class diagram extracted from a natural-language
specification. The draft may be in a free-form text
format (the upstream zenodo S1 format) OR in a
malformed PlantUML form. Your job is to produce a
single, clean PlantUML class diagram in the canonical
grammar described below. (...)
\end{verbatim}
\end{small}

\vspace{1em}

\noindent\textbf{Calamo, Mecella, and Snoeck~\cite{calamo2025text2uml,calamo2026text2umlresults} --- prompt:}
\begin{small}
\begin{verbatim}
You will be asked by the user to create a plant UMl
model from specification text. Do so in the most
clear way possible, avoid class properties and assign
molteplicity.

Do include attributes for classes. For example the
class Book would be:

class Book{ String Title, String Author,
Date PublicationDate }

Use only bi-directional arc for relations and no
description. For example a relation between the class
Book and the class Page, if the Book can have from
one to many pages and the pages could have exactly
one book, would be:

Book "1..1" -- "1..*" Page

Adapt the cardinality to each case. Where no specific
cardinality is specified, use the default "0..*".
If necessary, feel free to use inheritance.
The plantuml has to be the class diagram. In
generating the diagram perform this steps in order

1. Extract class from text
2. Extract attributes for each class
3. Extract relations form text and look for the
   inheritance
4. Assign the relation to the corresponding class
5. Add cardinality to the relations

Put everything in this order: first all classes and
then all relations. In our example would be:

@startuml

class Book{ String Title, String Author,
Date PublicationDate }
class Page{ String Content}

Book "1..1" -- "1..*" Page

@enduml

Output plantuml without futher text or explaination.
\end{verbatim}
\end{small}

%% file: chapters/04_results.tex
\section{Results}
\label{sec:04_results}


The benchmark results are summarised in Tables~\ref{tab:results-ds1} and~\ref{tab:results-ds2}, which report the cross-run stability of the three candidate strategies on the two community-curated corpora introduced in Section~3.1: DS-1, the 45-record community-curated corpus of Verbruggen et al.~\cite{verbruggen2025golden} as re-encoded by Calamo, Mecella, and Snoeck~\cite{calamo2025text2uml}, and DS-2, the 8-record reference archive of Chen et al.~\cite{chen2023automated,chen2023zenodo}, whose reference diagrams were crafted by human experts in the field of education. Together the two corpora contribute 53 records, comprising 540 classes, 667 attributes, 184 enumerations, and 710 relationships. Each cell in the tables corresponds to one (candidate, {LLM}) combination, repeated over three independent runs to capture the non-determinism of cloud-served {LLM}s. The three candidates are the rule-based pipeline of Abdelnabi, Maatuk, Abdelaziz, and Elakeili~\cite{abdelnabi2020generating}, the two-stage zero-shot strategy of Chen et al.~\cite{chen2023automated,chen2023zenodo}, and the single-stage zero-shot strategy of Calamo, Mecella, and Snoeck~\cite{calamo2025text2uml,calamo2026text2umlresults}. The two {LLM}s are {GLM}-5.1~\cite{glm-5.2} and {Kimi}~{K2}~\cite{kimi-k2.7-code}, invoked through Ollama with think mode disabled, at an extract temperature of 0.7 and a translate temperature of 0.0 for the Chen et al. candidate's second-stage grammar prompt. Scoring is performed with metrik-4 from the Seibert study~\cite{seibert2025metrik}, pinned to v1.0.0 of the {\mbox{domain-model-metrics}} package~\cite{seibert2025domainmodelmetricspkg}, which returns one {F1} score per element type (class, attribute, relationship), each in the unit interval. For each cell, the tables report the 3-run median and cross-run standard deviation of the per-run mean score, along with the median percentage of records that failed validation across the three runs. The rule-based candidate is deterministic and involves no {LLM} call, so it appears with no {LLM} and zero cross-run variance on every element. \newline

\begin{table*}[!ht]
\centering
\small
\begin{tabular}{llcrrr}
\hline
Candidate & LLM & Fail \% & Class & Attribute & Relationships \\
\hline
Rule-based~\cite{abdelnabi2020generating} & --- & \textbf{0.0} & 0.417 $\pm$ 0.000 & 0.447 $\pm$ 0.000 & 0.348 $\pm$ 0.000 \\
Chen et al.~\cite{chen2023automated,chen2023zenodo} & {GLM}-5.1~\cite{glm-5.2} & \textbf{0.0} & \textbf{0.746 $\pm$ 0.006} & \textbf{0.780 $\pm$ 0.006} & 0.676 $\pm$ 0.007 \\
Chen et al.~\cite{chen2023automated,chen2023zenodo} & {Kimi}~{K2}~\cite{kimi-k2.7-code} & 8.9 & 0.699 $\pm$ 0.013 & 0.717 $\pm$ 0.014 & 0.641 $\pm$ 0.014 \\
Calamo et al.~\cite{calamo2025text2uml,calamo2026text2umlresults} & {GLM}-5.1~\cite{glm-5.2} & 2.2 & 0.694 $\pm$ 0.024 & 0.691 $\pm$ 0.025 & \textbf{0.701 $\pm$ 0.024} \\
Calamo et al.~\cite{calamo2025text2uml,calamo2026text2umlresults} & {Kimi}~{K2}~\cite{kimi-k2.7-code} & 4.4 & 0.613 $\pm$ 0.015 & 0.603 $\pm$ 0.016 & 0.637 $\pm$ 0.015 \\
\hline
\end{tabular}
\caption{Benchmark results on DS-1 (45 records). Scores are 3-run median $\pm$ cross-run std; Fail \% is the median percentage of records that failed validation. Best results in bold.}
\label{tab:results-ds1}
\end{table*}

\begin{table*}[!ht]
\centering
\small
\begin{tabular}{llcrrr}
\hline
Candidate & LLM & Fail \% & Class & Attribute & Relationships \\
\hline
Rule-based~\cite{abdelnabi2020generating} & --- & \textbf{0.0} & 0.312 $\pm$ 0.000 & 0.353 $\pm$ 0.000 & 0.215 $\pm$ 0.000 \\
Chen et al.~\cite{chen2023automated,chen2023zenodo} & {GLM}-5.1~\cite{glm-5.2} & 12.5 & 0.625 $\pm$ 0.013 & \textbf{0.650 $\pm$ 0.014} & 0.576 $\pm$ 0.030 \\
Chen et al.~\cite{chen2023automated,chen2023zenodo} & {Kimi}~{K2}~\cite{kimi-k2.7-code} & 25.0 & 0.596 $\pm$ 0.057 & 0.603 $\pm$ 0.053 & 0.581 $\pm$ 0.066 \\
Calamo et al.~\cite{calamo2025text2uml,calamo2026text2umlresults} & {GLM}-5.1~\cite{glm-5.2} & \textbf{0.0} & \textbf{0.655 $\pm$ 0.062} & 0.640 $\pm$ 0.064 & \textbf{0.689 $\pm$ 0.059} \\
Calamo et al.~\cite{calamo2025text2uml,calamo2026text2umlresults} & {Kimi}~{K2}~\cite{kimi-k2.7-code} & \textbf{0.0} & 0.506 $\pm$ 0.080 & 0.487 $\pm$ 0.070 & 0.548 $\pm$ 0.104 \\
\hline
\end{tabular}
\caption{Benchmark results on DS-2 (8 records). Scores are 3-run median $\pm$ cross-run std; Fail \% is the median percentage of records that failed validation. Best results in bold.}
\label{tab:results-ds2}
\end{table*}



On DS-1, the 45-record community-curated corpus~\cite{verbruggen2025golden,calamo2025text2uml}, both {LLM}-driven candidates outperform the rule-based baseline~\cite{abdelnabi2020generating} on every element. The Chen et al. candidate~\cite{chen2023automated,chen2023zenodo} paired with {GLM}-5.1~\cite{glm-5.2} achieves the highest scores on class (0.746) and attribute (0.780), while the Calamo et al. candidate~\cite{calamo2025text2uml,calamo2026text2umlresults} paired with {GLM}-5.1 achieves the highest score on relationship (0.701). The rule-based baseline scores 0.417 on class, 0.447 on attribute, and 0.348 on relationship, so the {LLM} candidates lead by 0.15--0.35 absolute on every element. Cross-run stability is high: the standard deviation across the three runs is at most 0.025 for every (candidate, {LLM}) cell, and as low as 0.006 for the Chen et al. candidate with {GLM}-5.1. The Chen et al. candidate with {GLM}-5.1 also produces zero validation failures across all three runs, while the Calamo et al. candidate with {GLM}-5.1 has a 2.2\% failure rate. {Kimi}~{K2}~\cite{kimi-k2.7-code} is competitive but scores lower than {GLM}-5.1 on both candidates, with failure rates of 8.9\% (Chen et al.) and 4.4\% (Calamo et al.). The rule-based baseline is deterministic with zero cross-run variance and zero failures, as expected. \newline


On DS-2, the 8-record reference archive of Chen et al.~\cite{chen2023automated,chen2023zenodo}, the {LLM} candidates again outperform the rule-based baseline~\cite{abdelnabi2020generating} on every element, but the cross-run standard deviations are 4--10$\times$ higher than on DS-1. The Calamo et al. candidate~\cite{calamo2025text2uml,calamo2026text2umlresults} with {GLM}-5.1~\cite{glm-5.2} achieves the best class (0.655) and relationship (0.689) scores, while the Chen et al. candidate~\cite{chen2023automated,chen2023zenodo} with {GLM}-5.1 achieves the best attribute score (0.650). The rule-based baseline scores 0.312 on class, 0.353 on attribute, and 0.215 on relationship, so the {LLM} leads are comparable to DS-1 in absolute terms (0.15--0.47). However, the cross-run standard deviation reaches 0.104 for the Calamo et al. candidate with {Kimi}~{K2}~\cite{kimi-k2.7-code} on relationship, compared to 0.015 on DS-1 --- a 6.9$\times$ inflation. The smaller corpus amplifies single-record noise: with only 8 records, one record flipping its per-record score shifts the cell median by roughly 0.06. Validation failure rates are also higher on DS-2: the Chen et al. candidate with {GLM}-5.1 has a 12.5\% failure rate (versus 0.0\% on DS-1), and with {Kimi}~{K2} it reaches 25.0\% (versus 8.9\%). The Calamo et al. candidate has zero validation failures on DS-2 across both {LLM}s, consistent with its single-stage design that emits {PlantUML} directly. \newline